\documentclass[letterpaper, 10 pt, journal, twoside]{ieeetran} % 

\IEEEoverridecommandlockouts   

\usepackage{cite}
\usepackage{amsmath,amssymb,amsfonts}
\usepackage{algorithmic}
\usepackage{graphicx}
\usepackage{array,multirow,graphicx}
\usepackage{textcomp}
\usepackage{xcolor}
\usepackage{soul}

\usepackage{tabularx,booktabs}
\newcolumntype{C}{>{\centering\arraybackslash}X} % centered version of "X" type
\def\BibTeX{{\rm B\kern-.05em{\sc i\kern-.025em b}\kern-.08em
    T\kern-.1667em\lower.7ex\hbox{E}\kern-.125emX}}

\title{Bayesian and Neural Inference on LSTM-based Object Recognition from Tactile and Kinesthetic Information}

\author{Francisco~Pastor$^1$, Jorge~García-González$^2$, Juan M.~Gandarias$^1$,~\IEEEmembership{Student Member,~IEEE},
\\Daniel~Medina$^3$,~\IEEEmembership{Student Member,~IEEE}, Pau Closas$^4$,~\IEEEmembership{Senior~Member,~IEEE},\\ Alfonso~J.~García-Cerezo$^1$,~\IEEEmembership{Member,~IEEE}, Jesús M.~Gómez-de-Gabriel$^1$,~\IEEEmembership{Member,~IEEE}
\thanks{Manuscript received: June, 8, 2020; Revised August, 14, 2020; Accepted October, 26, 2020.}
\thanks{This paper was recommended for publication by Editor Tamim Asfour upon evaluation of the Associate Editor and Reviewers' comments.}
\thanks{This work is supported by the Spanish projects DPI2015-65186-R, RTI2018-093421-B-I00, the University of M\'alaga under grant UMA-CEIATECH-23, and the European Commission under grant agreement BES-2016-078237. This work has been partially supported by the NSF under Awards CNS-1815349 and
ECCS-1845833.}%
\thanks{$^1$F.Pastor, J.M. Gandarias, A.J.~García-Cerezo, and J.M. G\'omez-de-Gabriel are with the Robotics and Mechatronics Group, University of M\'alaga, Spain. {\tt\small \{fpastor, jmgandarias, ajgarcia, jesus.gomez\}@uma.es}}%
\thanks{$^{2}$J. Garc\'ia-Gonz\'alez is with the Department of Computer Languages and Computer Science, University of M\'alaga, Spain {\tt\small jorgegarcia@lcc.uma.es }}%
\thanks{$^{3}$D. Medina is with the Institute of Communications and Navigation, German Aerospace Center (DLR), Germany {\tt\small Daniel.AriasMedina@dlr.de}}%
\thanks{$^{4}$P. Closas is with the Department of Electrical and Computer Engineering, Northeastern University, Boston, MA, 02115, USA {\tt\small  closas@ece.neu.edu}}%
\thanks{Digital Object Identifier (DOI): see top of this page.}
}

\begin{document}

\maketitle

\markboth{This is a preprint. The original paper has been published in IEEE ROBOTICS AND AUTOMATION LETTERS. DOI: 10.1109/LRA.2020.3038377}
{Pastor \MakeLowercase{\textit{et al.}}: Bayesian and Neural Inference on LSTM-based Object Recognition from Tactile and Kinesthetic Information} 

\begin{abstract}
Recent advances in the field of intelligent robotic manipulation pursue providing robotic hands with touch sensitivity. 
Haptic perception encompasses the sensing modalities encountered in the sense of touch (e.g., tactile and kinesthetic sensations). 
This letter focuses on multimodal object recognition and proposes analytical and data-driven methodologies
to fuse tactile- and kinesthetic-based classification results. The procedure is as follows: 
a three-finger actuated gripper with an integrated high-resolution tactile sensor performs squeeze-and-release Exploratory Procedures (EPs).
The tactile images and kinesthetic information acquired using angular sensors on the finger joints constitute the time-series datasets of interest.
Each temporal dataset is fed to a Long Short-term Memory (LSTM) Neural Network, which is trained to classify in-hand objects.
The LSTMs provide an estimation of the posterior probability of each object given the corresponding measurements, which after fusion allows to estimate the object through Bayesian and Neural inference approaches.
An experiment with 36-classes is carried out to evaluate and compare the performance of the fused, tactile, and kinesthetic perception systems.The results show that the Bayesian-based classifiers improves capabilities for object recognition and outperforms the Neural-based approach.
\end{abstract}

\begin{figure}
    \centering
    \includegraphics[width = 0.78\columnwidth]{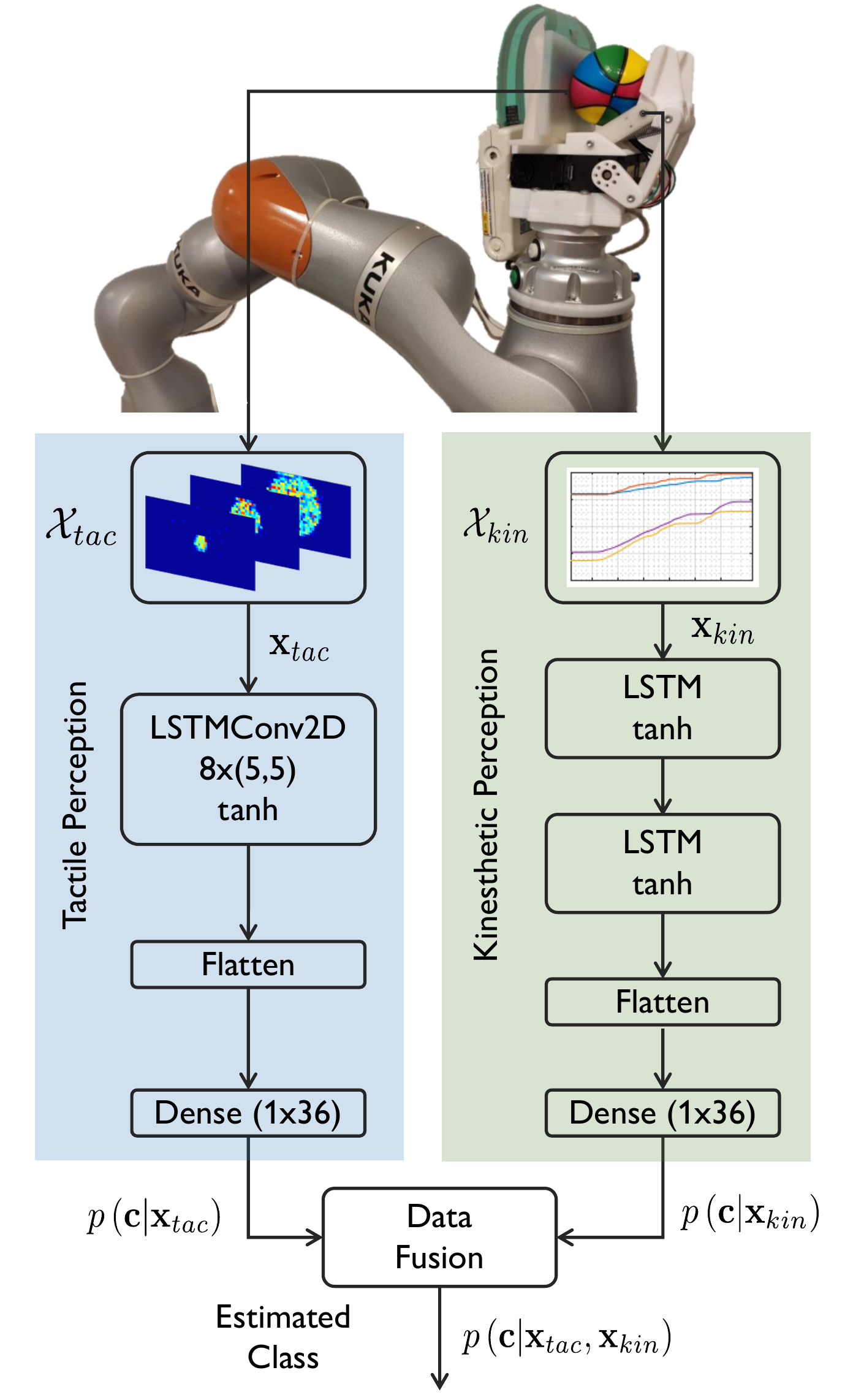}
    \caption{Representative schematic of the proposed methodology. Sequences of tactile images feed the tactile network, and the time-series of the joint angles of the gripper feed the kinesthetic network. The activation of the last layers of each network are fused to output an estimated output class.}
    \label{fig:representative}
\end{figure}

\section{Introduction}

\IEEEPARstart{T}{ouch} sense is essential for human beings. One commonly refers to touch sensing as the sensations perceived through the skin, including tactile and kinesthetic sensations. Kinesthetic data is perceived as two sources of information, the data received from the relative position of limbs and the dynamical forces exerted by the muscles. The combination of kinesthetic and tactile information defines the haptic perception terminology. We would not be able to carry out many vital activities without the sensations perceived in this sense. This requirement is also present in the field of robotics\cite{robles2006importance,Jamone2015HighlyHand, LuoShanBimboJoaoDahiyaRavinder2017RoboticReview}. Recent advances on intelligent robotic systems consider the integration of artificial touch sense to improve perception~\cite{Bartolozzi2016RobotsTouch, Gandarias2018EnhancingInteraction}.

Humans are capable of carrying out extremely complex manipulation tasks even without looking at the grasped objects. This illustrates our high capability for haptic recognition, which allows us to distinguish objects just by touch. Many of these tasks are yet impossible for the most advanced robots~\cite{Dahiya2010TactileHumanoids}.
Nonetheless, multiple robotic applications benefit from autonomous manipulation, such as search-and-rescue~\cite{Gandarias2018TactileInDisasterScenarios}, healthcare~\cite{Trujillo-Leon2018TactileWheelchairs}, or~\cite{aggarwal2014object}. In essence, haptic perception results integral for those applications where vision is compromised due to inadequate light conditions, absence of textures or variety of colors.

This letter tackles the problem of haptic object recognition considering the fusion of temporal tactile and kinesthetic information. In particular, LSTMs-based models to process the time-series of haptic data are proposed. It results intuitive thinking that the fusion of tactile and kinesthetic information will lead to an enhanced object recognition process, thus limiting the shortcomings arising from the single sensing modalities. The performance of multisensor inference is addressed using both \textit{classical} (i.e., analytic, deterministic or Bayesian) and \textit{data-driven} (i.e., machine learning- or neural-based) inference.
A Bayesian fusion strategy is presented, for which the discrete probabilities for classification estimated by the pertinent LSTMs participate in a maximum a posteriori (MAP) estimation problem. The optimal fusion rule corresponds to the joint posterior distribution derived from the MAP estimator. Besides, a data-driven fusion approach, in which a fully-connected layer learns to fuse discrete probabilities estimated by the pertinent LSTMs is also presented. A representative schema of our methodology is shown in Fig.~\ref{fig:representative}. In particular, the main contributions of this work are:
\begin{itemize}
    \item A LTSM-based methodology for haptic object recognition via tactile and kinesthetic sensing modalities.
    \item Formulation of multimodal data fusion for classification based on  Bayesian and Neural-based approaches.
    \item Integration and evaluation of the proposed methodology with real data collected from a sensorized, three-finger underactuated gripper.
\end{itemize}
The performance of the proposed methodology is evaluated in a 36-classes experiment, including a comparison between the Bayesian and Neural fusion approaches, and the tactile and kinesthetic perception systems with reduced data redundancy. The code and dataset are publicly available in a GitHub repository \footnote{https://github.com/fpastorm/LSTM-Haptic-Fusion}.

This paper is organized as follows: Section~\ref{sec:state_of_the_art} presents the related work and the state-of-art in object recognition. Section~\ref{sec:haptic_perception} details the problem of haptic perception for object recognition. In Section~\ref{sec:Fusion}, the methods for Bayesian and Neural data fusion for classification problems are described. The experimental protocol is presented Section~\ref{sec:experimental_protocol}, followed by the discussion on the obtained results in Section~\ref{sec:results and discussion}. Finally, Section~\ref{sec:conclusions} presents the conclusions and prospective research work.

\section{Related Work}
\label{sec:state_of_the_art}

\subsection{Surface and In-hand Object Recognition}

The problem of tactile-based object recognition has been studied in a variety of works.
One common approach relates to surface or material discrimination~\cite{Kaboli2018Robust}. A system for the estimation of surface friction is proposed in~\cite{jamali2011majority}. The system uses machine learning to distinguish textures detected by a bio-inspired artificial finger, achieving recognition rates around $95\%$. A haptic EP is presented in~\cite{liu2012surface} for recognizing object surfaces with an intelligent finger. Supervised learning algorithms are evaluated and compared achieving classification accuracies of $88.5\%$. GelSight, a camera-based tactile sensor,  is another convenient option for this problem, allowing to deal with tactile textures as images~\cite{li2013sensing}.
However, surface recognition approaches would not be a good strategy for in-hand object recognition, especially when dealing with different objects made of the same material. For this reason, other tactile properties such as stiffness or shape are reliable sources of information for in-hand manipulation tasks.

Many of the haptic perception approaches solely exploit static tactile information. Traditional computer vision and machine learning methods have been used to face the problem of tactile object recognition~\cite{Luo2015NovelRecognition}. The work presented in~\cite{li2014localization} uses a feature-based matching technique from tactile images obtained with GelSight, while deep learning algorithms were proposed in~\cite{gandarias2019cnn, pastor2019using}. 
Nevertheless, other researchers faced the tactile object recognition problem by considering dynamic tactile information.  
Some studies have followed this approach with different methodologies~\cite{Madry2014ST-HMP:Data, Liu2016ObjectMethods, gandarias2019active}. 
A tactile-based method to identify the Center of Mass of rigid objects for object discrimination is presented in~\cite{yao2017tactile}.
Long Short-term Memory (LSTM) Neural Networks have proven excellent results when dealing with sequential tactile data~\cite{Zapata-Impata2019LearningDetection}.

Transfer learning is a recent alternative option explored for computer vision applied to recognition, as in ~\cite{kaboli2016re, feng2018active, kaboli2018active, gandarias2019cnn}.

\begin{table}
\centering
\caption{State-of-art on multimodal perception with haptic properties (V refers to Vision).}
\label{tab: state_art}
\begin{tabularx}{0.45\textwidth}{@{}c*{5}{cll}c@{}}
\toprule
Year/Work  & Approach & Properties \\ 
\midrule
2011 /~\cite{kroemer2011learning} & Data-driven & V+Audio \\
2016 /~\cite{Gao2016DeepData}  & Data-driven & V+Pressure+Vibration+Temperature \\
2017 /~\cite{Liu2017VisualTactileRecognition} & Analytical & V+Pressure Images\\
2017 /~\cite{Falco2017Cross-modalExploration} & Data-driven & V+Pressure Images\\
2018 /~\cite{Pfanne2018FusingEstimation}  & Analytical & V+Proprioception  \\
2019 /~\cite{lee2019making} & Data-driven & V+Force+Proprioception \\

\midrule
2013 /~\cite{xu2013tactile} & Analytical & Pressure+Vibration+Temperature \\
2017 /~\cite{kaboli2017tactile} & Data-driven & Temperature+Vibration+Force \\
2018 / ~\cite{GomezEguiluz2018MultimodalSensing}  & Analytical & Temperature+Vibration \\
2019 /~\cite{luo2019iclap} & Analytical & Pressure Images+Proprioception \\
2020 / [Ours] & Hybrid & Pressure Images+Proprioception\\ 
\bottomrule
\end{tabularx}
\end{table}

\begin{figure*}[ht!]
    \centering
    \includegraphics[width =1\textwidth]{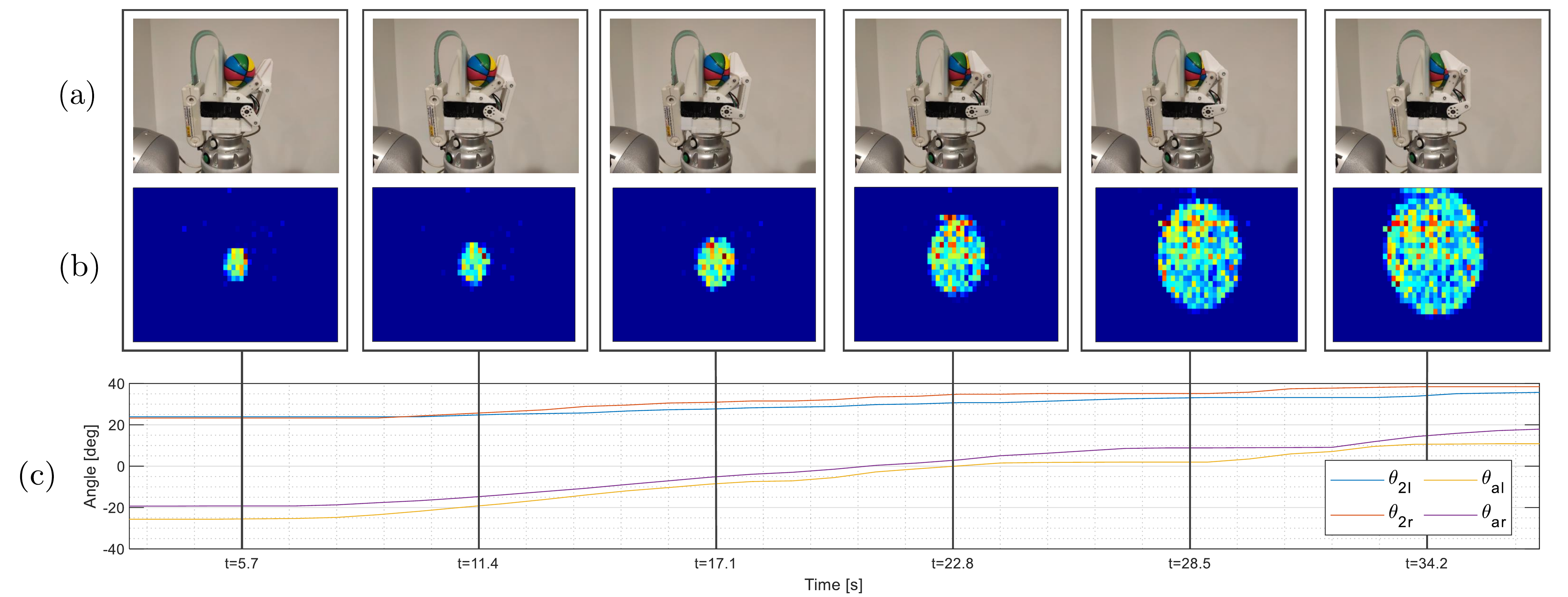}
    \caption{Excerpt of time sequences of tactile and kinesthetic data during the EP~(a). Tactile information~(b) consist of series of tactile images. Note that in this figure, six tactile images are represented to illustrate the variation of the pressure distribution along the sequence. Kinesthetic information~(c) is formed by the variation of the joints position of the underactuated fingers. Here, $\theta_al$ and $\theta_ar$ refer to the actuated joint angle of the left and right fingers, respectively; and $\theta_2l$ and $\theta_2r$ refer to the underactuated joint angle of the left and right fingers, respectively. A more detailed description of the underactuated gripper is presented in section~\ref{sec:experimental_protocol}.}
    \label{fig:data_collection}
\end{figure*}

\subsection{Tactile-based Object Recognition via Multiple Physical Properties}
Multimodal perception is an important research topic in robotics. Fusing multiple sources of information provides excellent advantages to tackle perception challenges~\cite{araki2011autonomous, navarro2012haptic,han2020hands}, and has been extensively used in multiple domains with excellent results~\cite{Library2017OnlineRecognition,chou2018mmdf,medina2019gnss}. 
Recent trends consider the problem of integrating multiple haptic or tactile-based sources of information. Table~\ref{tab: state_art} summarizes the state-of-art with reference to multimodal haptic perception, including the methodology proposed in this work. One of the most common approaches is based on fusing tactile and visual data~\cite{Gao2016DeepData, Liu2017VisualTactileRecognition, lee2019making}. 
The joint use of tactile frequencies recorded with a microphone in contact with the surface and visual images of the objects for dynamic surface discrimination was introduced in~\cite{kroemer2011learning}.  
The fusion of joint positions and visual information for estimating the pose of the in-hand object is presented in~\cite{Pfanne2018FusingEstimation}. A cross-modal framework for visuotactile object recognition is proposed in~\cite{Falco2017Cross-modalExploration}. Nevertheless, these methods might not be a good strategy in those scenarios where vision is compromised.

Despite the advantages of combining multiple haptic-based sources for object recognition, just a few research studies have followed this approach. Kaboli et al. proposed a tactile-based framework for autonomous exploration and object recognition based on physical properties~\cite{kaboli2017tactile}. A multimodal tactile sensor (BioTac~\textregistered) was employed in~\cite{fishel2012bayesian, xu2013tactile} to identify objects by their compliance, texture, and thermal properties.
Luo et al. proposed a method to synthesize kinesthetic and tactile information~\cite{luo2019iclap}. 3D positions and the pressure value of each tactel of a sensor formed a 4D-point cloud that feeds a classification algorithm. The combination of proprioceptive and tactile data to improve the accuracy in object identification tasks is presented in~\cite{Schiefer2018ArtificialTasks}. A recursive tactile sensing approach is proposed in~\cite{GomezEguiluz2018MultimodalSensing} for multimodal material recognition. In general, standard classifiers (e.g., SVMs, Neural Networks) output a probabilistic distribution where each class gets a probability that can be easily exploited by Bayesian and Neural approaches. %

\section{Haptic Perception for Object Recognition}
\label{sec:haptic_perception}

There are multiple ways of carrying out an Exploratory Procedure (EP) to obtain different haptic information~\cite{Okamura2001FeatureFingers}. In the particular case of discerning in-hand objects, stiffness and shape are two of the most relevant haptic features. A natural EP to perceive information about the stiffness of in-hand objects is to palpate them during a squeeze-and-release process. Besides, kinesthetic information of the fingers' position during this EP provides data about shape. Fig.~\ref{fig:data_collection}~(a) shows an excerpt from the EP followed in this work.

Similarly, there are many ways of measuring, representing, and processing tactile data. On the one hand, pressure images are collected from a tactile sensor formed by a pressure sensor array. The details of the sensor are described in section~\ref{sec:experimental_protocol}. Tactile data for each grasp are represented as sequences of tactile images (see Fig.~\ref{fig:data_collection}~(b)). 
On the other hand, kinesthetic perception usually relates to the information perceived through the muscles or joints of the body. In particular, kinesthetic sensations allow us to know the dynamic spatial configuration of our body (relative position) and the dynamical efforts of our muscles (forces). In this work, only the first type of kinesthetic information is used.
As represented in Fig~\ref{fig:data_collection}~(c), the dynamic kinesthetic information is represented as time series of the joint angles of the fingers ($\theta_al$, $\theta_ar$, $\theta_2l$, and $\theta_2r$). The subindex $a$ refers to the actuator joint, the sub-index $2$ refers to the second joint of the finger (in this case, the underactuated joint), and the sub-indices $r$ and $l$ refer to the right and left fingers, respectively.

As represented in Fig.~\ref{fig:representative}, the core element within both tactile and kinesthetic models is the LSTM layer. 
Long Short-term Memory networks~\cite{Hochreiter:1997} are a special kind of Recurrent Neural Network (RNN) that adds long-term memory to RNN by allowing a constant error back-propagation within their inner memory cells. This fact makes them a very good choice to process information with a robust temporal structure, even if the temporal relation is not immediate, and their use is prevalent in predictive~\cite{zyner2018recurrent} and classification~\cite{park2018multimodal} problems.

In this work, sequences of tactile images are used to train the tactile model. This model uses only one ConvLSTM~\cite{shi:2015} layer.
The ConvLSTM layer is followed by a fully-connected layer with 36 neurons (which match the number of classes of the experiment presented in section~\ref{sec:experimental_protocol}), and a Softmax function to provide the output classification distribution ($p\left( \mathbf{c} | \mathbf{x}_{tac} \right)$).
The kinesthetic model, on the other hand, uses two LSTMs~\cite{Hochreiter:1997} to allow a progressive codification of its vector to shape temporal data. The second LTSM is followed by a fully-connected layer with 36 neurons and a Softmax function to provide the output classification probability distribution ($p\left( \mathbf{c} | \mathbf{x}_{kin} \right)$).

After training the tactile and kinesthetic models, the classification outputs from each model when classifying an unseen object can differ depending on the input data and the training process. Hence, to achieve an accurate and robust classification performance, the outputs from each model are fused in an analytical and a data-drive classification approach. The next section describes the proposed fusion methodologies.

\section{Bayesian and Neural Inference Methodology}

\subsection{Bayesian Data Fusion for Probabilistic Classification}
\label{sec:Fusion}

Let us define $\mathbf{x}_{i}\in\mathcal{X}_i$ as the observations from a particular set upon which the classification problem is resolved (i.e., the tactile images $\mathbf{x}_{tac}$ and the kinesthetic information $\mathbf{x}_{kin}$). The class labels ${\mathbf{c}\in\mathcal{Y}}$ are defined prior to the training, and the total number of labels is denoted with $N$. Let $\mathcal{H}$ be the family of classifiers considered (i.e., the output of afore-described LSTMs, considering haptic and kinesthetic sensing respectively). A classifier behaves as mapping function between the sample and the label spaces, ${h\in\mathcal{H}, h:\mathcal{X}\mapsto\mathcal{Y}}$. In general, classifiers can be categorized as \textit{ordinaries} or \textit{probabilistic}~\cite{friedman2001elements}. This work focuses on the latter, with a classifier estimating 
a conditional discrete probability function, such that
each class is assigned a correspondence probability~\cite{arribas2005model}.

Thus, we obtain posterior distributions $p\left(\mathbf{c}|\mathbf{x}_{tac}\right), p\left(\mathbf{c}|\mathbf{x}_{kin}\right)$ from the LSTMs using tactile and kinesthetic data respectively. To ease the notation, hereinafter we refer to tactile and kinesthetic data as $\mathbf{x}_1$ and $\mathbf{x}_2$ respectively. The classifier posterior distributions can be then modeled as categorical
\begin{equation}
    p\left(\mathbf{c}|\mathbf{x}_i\right) = \prod_{j=1}^N p_{j|i}^{[c=j]},
\end{equation}
where $p_{j|i}$ denotes the probability for the $j$-th class given the $i$-th classifier, and the Iverson bracket $[c=j]$ is an indicator function that returns $1$ if $c=i$ and $0$ otherwise. 
The a priori class probability $p\left( \mathbf{c}\right)$ is categorical as well and is defined by the probabilities $p_{j|0}$, which in the equiprobable case result in $p_{j|0}=1/N$, $\forall j$. 
The optimal fusion rule is provided by the joint posterior distribution as
\begin{align}
    \notag p\left( \mathbf{c} | \mathbf{x}_{1}, \mathbf{x}_{2} \right) &= \frac{p\left( \mathbf{x}_{2} | \mathbf{c}, \mathbf{x}_{1} \right)\cdot p\left( \mathbf{c}|\mathbf{x}_{1} \right)}{p\left( \mathbf{x}_{2}|\mathbf{x}_{1} \right)} \\
    \notag & \propto %\myeq \  
    \frac{p\left( \mathbf{c} | \mathbf{x}_{2} \right) \cdot p\left( \mathbf{c} | \mathbf{x}_{1} \right)   }{ p(\mathbf{c}) } \\
    &= \prod_{j=1}^N {\Big( \frac{ p_{j|1}\cdot p_{j|2} }{ \underbrace{p_{j|0}}_{\text{prior class prob.}} } \Big)}^{[c=j]} \;,
\end{align}
\noindent where we have used that $\mathbf{x}_{2}$ and $\mathbf{x}_{1}$ are conditionally independent given $\mathbf{c}$. The resulting joint distribution is again modeled as categorical random variable from which the mode (i.e., the estimated class) can be easily found as the class with the MAP probability. As a consequence of using MAP estimation, computation of the normalizing constant $p\left( \mathbf{x}_{2}|\mathbf{x}_{1} \right) / p\left( \mathbf{x}_{2}\right)$ is not required in the joint distribution, which can be easily obtained by normalization.

\begin{figure}
    \centering
    \includegraphics[width=1\columnwidth]{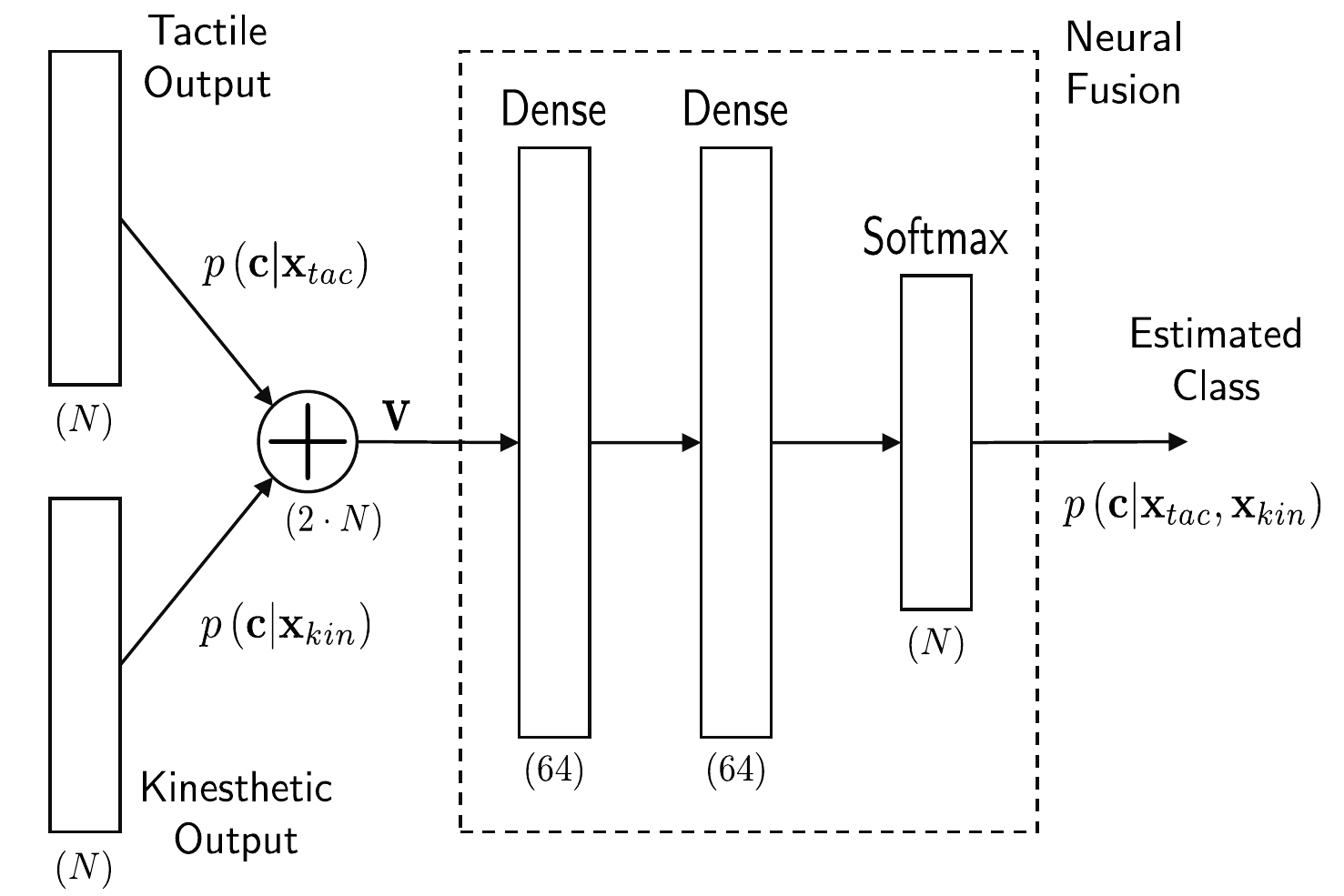}
    \caption{Schematic illustration of the data-driven inference method. Tactile and Kinesthetic outputs are concatenated and fed to a two-layer fully-connected network that learns to fuse the probability distributions from tactile and kinesthetic classifiers. The numbers in parentheses establish the length of the input and output vectors and the number of neurons of the dense layers.}
    \label{fig:perceptron}
\end{figure}

\subsection{Neural-driven Data Fusion for Classification}

A data-driven approach leverages on supervised ML algorithms to imitate (learn) fusion rules based on a set of training data. Such rules would, under an asymptotic regime or when fed large volumes of data, recreate the statistical inference basis, as well as some \textit{hidden} rules directly related to the problem of interest. This approach can be adopted in two ways: i) training an entire multichannel network at once~\cite{Kerzel}; ii) training each channel's network separately, and then training a fusion layer~\cite{tatiya2019deep}. The latter is the solution adopted in this letter to compare the results of neuronal and Bayesian fusion. 

\begin{figure*}
    \centering
        \includegraphics[width=1
    \textwidth]{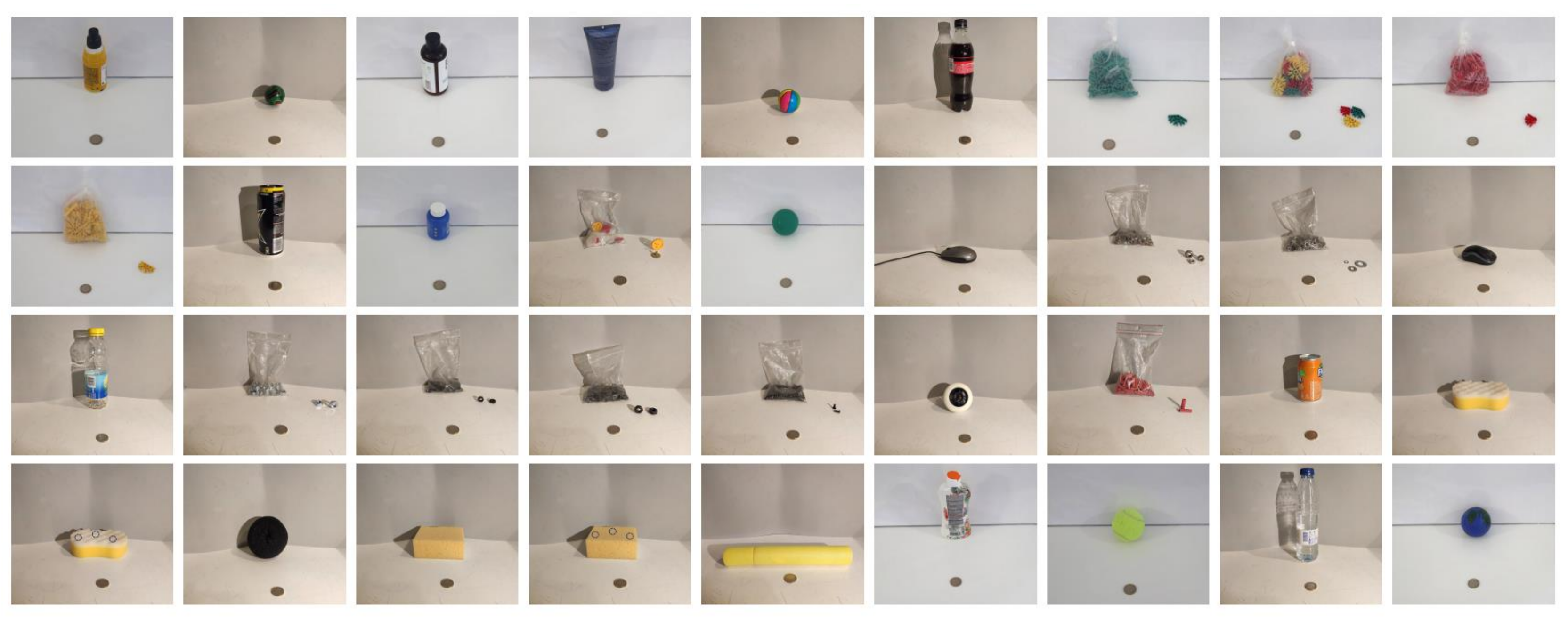}
    \caption{Pictures of the 36 objects used in experiments. From left to right:1.~antiseptic bottle, 2.~ball1, 3.~shampoo, 4.~body gel, 5.~ball2, 6.~bottle of coke, 7.~green bricks, 8.~mixed bricks, 9.~red bricks, 10.~yellow bricks, 11.~energy drink can, 12.~flux paste, 13.~gears, 14.~ball3, 15.~mouse1, 16.~mixed nuts, 17.~mixed washer, 18.~mouse2, 19.~bottle of ice tea, 20.~M10 nuts, 21.~M6 nuts, 22.~M8 nuts, 23.~rivets, 24.~skate wheel, 25.~rubber pipes, 26.~soda can, 27.~sponge rough, 28.~sponge rough with inclusions, 29.~sponge scrunchy, 30.~sponge soft, 31.~sponge soft with inclusions, 32.~sponge pipe, 33.~sunscreen, 34.~tennis ball, 35.~bottle of water, 36.~world ball.}
    \label{fig:objects}
\end{figure*}

Our neural solution consists of a two-layer fully-connected network which interprets the outputs from the tactile and kinesthetic networks as inputs, and learns to fuse the haptic data ---i.e., learns the data fusion axioms. A representative scheme of the method is depicted in Fig.~\ref{fig:perceptron}. The discrete probability distributions given by the tactile and kinesthetic classifiers, $p\left( \mathbf{c} | \mathbf{x}_{tac} \right)$ and $p\left( \mathbf{c} | \mathbf{x}_{kin} \right)$ respectively, are concatenated to define a single input vector $\mathbf{v}\in\mathbb{R}^{2 \cdot N}$, which feeds the fusion network. Then, two dense layers formed by 64 neurons learn to fuse the haptic data. These layers are followed by a softmax layer of 36 neurons that outputs the fused predicted class $p\left( \mathbf{c} | \mathbf{x}_{tac}, \mathbf{x}_{kin} \right)$, whose (normalized) distribution relates to a beta distribution.
Unlike the analytical method previously described, this method is based on supervised learning only. Therefore, the fusion rules to be learned will depend on the data set used to train the network. Thus, it is not possible to predict its performance until the method is trained and the experiments are conducted.

\section{Experimental Protocol}
\label{sec:experimental_protocol}

This section describes the design and mechanical components of the gripper, and the characteristics of the sensors. Besides, the data collection process followed to obtain the tactile and kinesthetic information is presented.

\subsection{Underactuated Gripper}

The experiments are developed using an underactuated gripper with three fingers (see Fig.~\ref{fig:representative}), which was mainly designed and 3D printed for this research. Two of the fingers are independently actuated with two phalanxes and two DOFs. These fingers employ a spring to keep a passive torque over the underactuated joint when no forces are applied. Two potentiometers (\emph{muRata} SV01 $10 k\Omega$ linear) are employed to measure the distal joint angles.  The gripper integrates two \emph{Dynamixel XL450-W250} servos featuring a digital magnetic encoder ($0.088^{\circ}$ resolution). In order to simplify the experimental setup, the motors integrate an open-loop force control, where the actuation (pulse-width modulation - PWM) of the direct current (DC) motors of the smart servos follow a slow triangular trajectory from a minimum value (5\%) to a maximum (90\%) of the maximum torque of $1.4~\textrm{N}\cdot\textrm{m}$ of each actuator. 
The third fixed finger defines the whole gripper. It has a planar surface that  holds the tactile sensor: a Tekscan sensor model 6077, with 1400 tactels arranged in a $28 \times 50$ matrix with a density of $27.6$  tactels/cm$ ^2$. The sensor also has a silicon pad covering, which protects the sensors, softens, and distributes the applied pressure.

\subsection{Data Collection}
The EP consists of a squeeze-and-release process, during which both kinesthetic and tactile sequences are collected from a particular object. The kinesthetic data consist of the angles between phalanxes $\theta_{2_r}, \theta_{2_l}$ and the joint angles of the actuator $\theta_{a_r}, \theta_{a_l}$, which result in a temporal array $\boldsymbol{\theta} \in \mathbb{R}^{4 \times \mathcal{K}}$, where $\mathcal{K} = 41$ is the number of samples of the sequence over time (see Fig.~\ref{fig:data_collection}). These sequences are measured using joint position sensors at the underactuated fingers. 
Tactile data, on the other hand, are collected with the tactile sensor of the fixed finger. Tactile sequences are composed of the variation of pressure over time $\mathcal{P}\in\mathbb{R}^{28 \times 50 \times \mathcal{T}}$, where $\mathcal{T} = 21$ is the number of samples of the sequence over time. 

\begin{figure}
    \centering
    \includegraphics[width=1 \columnwidth]{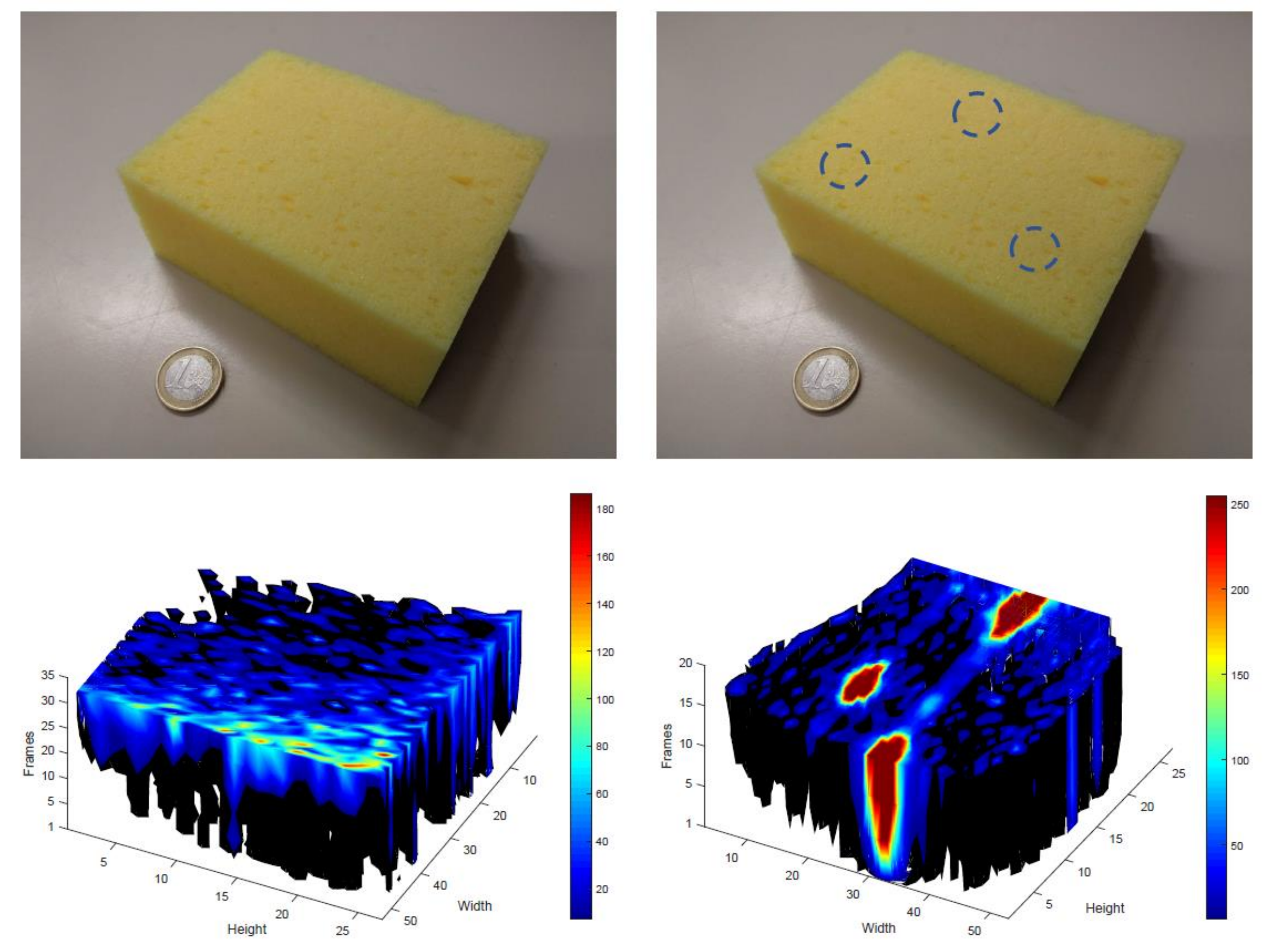} 
     \caption{Illustration of the effects of the hard inclusions on the tactile measurements. A deformable object (sponge) with and without inclusions (top), and their respective 3D tactile tensors (bottom).}
     \label{fig:representation_tactile}
\end{figure}

Overall, a dataset of 36 objects (depicted in Fig. \ref{fig:objects}) with 60 sets of tactile and kinesthetic data (60 grasps) for each object has been collected, forming a total of 1440 experiments.
To collect data with diverse, but related, tactile features, we can categorize the objects as rigid, soft, and in-bag objects. Therefore, objects from each category have similar properties, which makes the classification problem more challenging. Besides, hard inclusions (marbles) are inserted inside some deformable objects in order to determine how well the method can discriminate identical objects with different internal features (see Fig.~\ref{fig:representation_tactile}).
During the data collection, each object is manually located by a human operator with an arbitrary position and orientation to have random data for each grasp.

\subsection{Training}

With respect to the Bayesian-based inference approach, the tactile neural network classification model was trained on 15 examples for each class (25\% of the total dataset) using 20\% of the training set for validation purposes, over 30 epochs. Adam optimizer is used with a learning rate of $0.0001$ and Categorical Crossentropy as the loss function. The kinesthetic neural network classification model is trained on the same 15 examples for each class as the tactile network, using the same 20\% of the training set for validation over 700 epochs.
Adam optimizer is used with a learning rate of $0.00001$  and also with Categorical Crossentropy as the loss function. Furthermore, both tactile and kinesthetic data are normalized before being used for training and testing. The remaining 45 examples form the test set.

Regarding the data-driven approach, the tactile and kinesthetic neural networks classification models are trained as aforementioned but using ten examples from each class for training.
Then, the fusion neural network is trained using Adam optimizer with $0.0001$ learning rate over 200 epochs using five examples from each class. Hence, the same 45 examples remain for testing purposes to offer a proper comparison with the Bayesian Inference approach.

\section{Results and Discussion}
\label{sec:results and discussion}
The performance of the proposed methodology has been evaluated in terms of the recognition rate. To have meaningful statistical performance metrics, the training and testing procedures are repeated a total of 20 times.

\begin{figure*}
    \centering
    \includegraphics[width=1\textwidth]{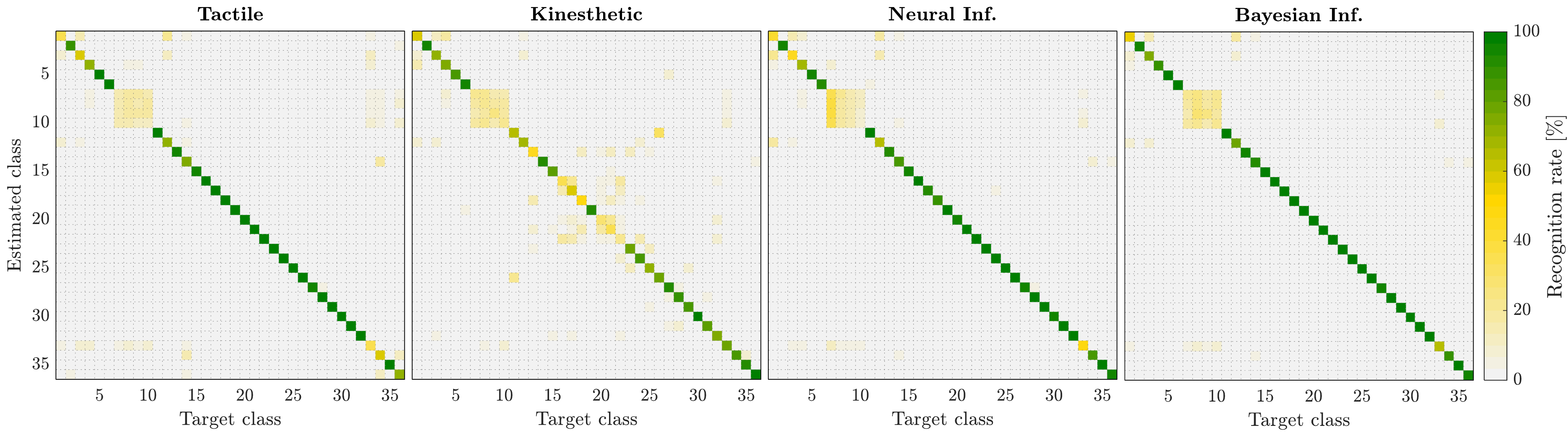}
    
    \caption{Confusion matrices (CM) for the classifiers, where the abscissa refers to the target or actual class, and the ordinate is the estimated class. The success or recognition rate is based on the color code depicted on the right. From left to right: i) the CM for the tactile-based classification; ii) the CM for the kinesthetic-based classification; iii) the classification resulting of the Neural fusion of the tactile- and kinesthetic-based classifiers; iv) the CM of the Bayesian inference on tactile- and kinesthetic-based classifiers.}
    \label{fig:confusion_matrices}
\end{figure*}

Each network has been trained and tested in each experiment with random and different training, validation, and test sets. One aspect to be considered at this point is that, although the sets of data were different in each experiment, the corresponding data from the same grasp feed the kinesthetic and tactile models. This means that each model was trained and tested with the data for the same grasps in each experiment. This aspect is essential for the proposed approach because the data that feed each model must come from the same grasp. The resulting mean Confusion Matrices~(CM) for each classifier (kinesthetic only, tactile only, and fusion-based) are represented in Fig.~\ref{fig:confusion_matrices}. These CMs represent the average results of each classifier for the whole set of experiments.

As shown in Fig.~\ref{fig:confusion_matrices}, the fusion-based approaches, with the highest success rate, outperforms the simple tactile and kinesthetic classifiers. In particular, the Bayesian-based method presents better performance than the Neural-based one. We can also see that the success rate is higher for the tactile classifier than for the kinesthetic one. This statement makes sense indeed as the sequences of tactile images provide more information about the shape and stiffness than the kinesthetic data. Another aspect to mention is that objects from 7 to 10 cannot be identified by any method. These objects are practically identical, as can be seen in Fig.~\ref{fig:objects}, and are not distinguishable by the sense of touch, not even by a human.

\begin{figure}
    \centering
    \includegraphics[width = 1 \columnwidth]{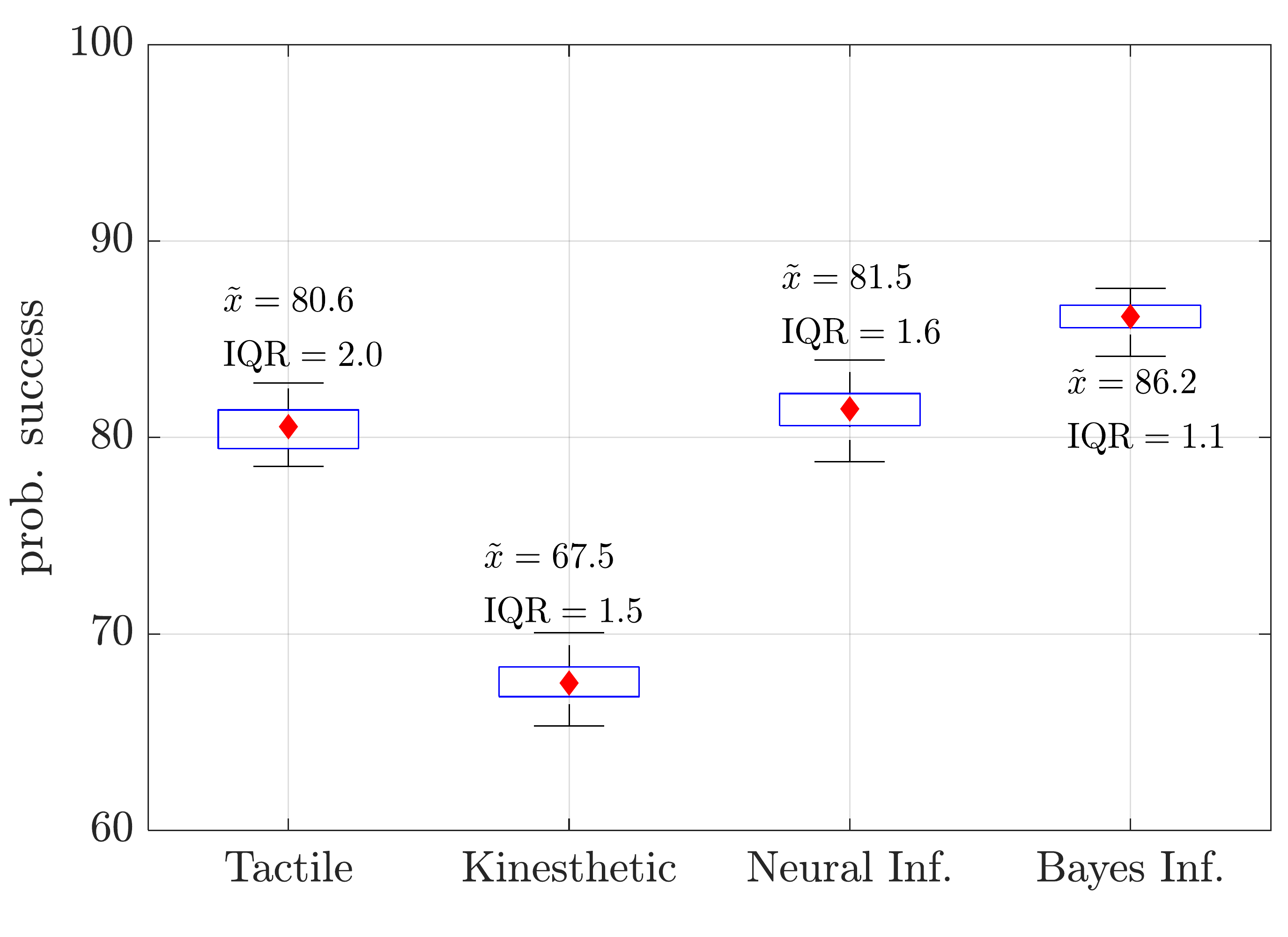}
    \caption{Representation and comparison of the probability distributions of success achieved by the tactile classifier, the kinesthetic classifier, the Neural, fusion-based classifier, and the Bayesian, fusion-based classifier.}
    \label{fig:boxplot}
\end{figure}

Fig.~\ref{fig:boxplot} presents a box plot for the recognition rates achieved by the four classifiers. The mean values are: i)~tactile model: 80.6\%; ii)~kinesthetic model: 67.5\%; iii)~Neural-based model: 81.5\%; iv)~Bayesian-based model:86.2. 
Thus, it is evident that the Bayesian-based method presents the highest recognition rate, while the tactile model achieves better performance than the kinesthetic one. However, the tactile probability distribution presents a higher standard deviation, which means that the tactile model presents a higher dependency towards the training data than kinesthetic-based modeling. Moreover, this outcome reveals the greater independence of the Bayesian-based model to the training data. 

On the one hand, regarding the comparison of the data-driven and analytical inference approaches, these results reflect that the Bayesian method outperforms the neural counterpart.
One reason being that the proposed Bayesian inference method is optimal for the fusion of discrete probability distributions. This method includes analytical rules for the fusion of this type of data. However, the neuronal inference method learns these rules from previous experiences, depending on the training data. Hence, for the same, small training set, the Bayesian approach achieves the best performance. 

On the other hand, the Neural-based approach could enhance its performance considerably when trained on an extensive training set. This method could potentially learn fusion rules not considered by the Bayesian method and that are more related to the fusion that we humans perform. This idea should be handled with caution and studied in depth in future work. Nevertheless, getting a large amount of tactile data is a complex task. This problem could be addressed by different techniques such as a sim-to-real approaches that uses simulated data to pre-train the models; or using Generative Adversarial Networks (GANs) or Variational Autoencoders to generate new data similar to that obtained by the real sensor. 
Comparing these outcomes with related works in the field of multimodal haptic perception is complex. The main reason is that each approach uses different sensing modalities and hardware, considering different properties and objects (see Table~\ref{tab: state_art}). Li et al. have recently published a review that covers the most relevant sensor types and solutions of the state-of-art in the field of tactile perception~\cite{li2020review}.

\section{Conclusions}
\label{sec:conclusions}

This letter proposed a methodology to fuse tactile and kinesthetic information for multimodal, haptic object recognition. A combination of data-driven and analytical methods based on LSTM, and Bayesian and Neural fusion models was presented.
A squeeze-and-release EP with a three-finger underactuated gripper was used to collect tactile and kinesthetic dynamic data from in-hand objects. The haptic data of 36 objects with different physical properties were acquired to form a dataset with training and testing purposes.
The outcomes of this work were two-folded: first, from an application point of view, temporal tactile and kinesthetic data are fed to LSTMs to perform the task of unimodal, haptic object recognition; second, from an estimation theory point of view, the performance of multi-sensor fusion classical inference based on a MAP estimator is compared to that of supervised learning-based inference based on a fully-connected network. 

The experimental results presented in this letter showed the benefits and potential of combining different haptic sources of information for multimodal haptic object recognition. In particular, the proposed methodology exhibits the advantages of combining advanced deep learning and Bayesian-based models for high-level and accurate haptic perception. Moreover, these results evidence that, for problems with reduced data redundancy, classical fusion rules outperform neural-based inference. Future research shall consider the creation of a benchmark to tackle the problem of comparing multimodal haptic-based approaches. Such a benchmark shall include a large variety of objects with different physical properties, recording data from multiple sensors to integrate other sources of haptic information such as inertial measurements, force, or temperature. Furthermore, the combination of the proposed multimodal haptic perception approach with vision systems will also be considered to enhance dexterous manipulation tasks.

\par
\bibliographystyle{IEEEtran}
\bibliography{biblio.bib}

\begin{thebibliography}{10}
\providecommand{\url}[1]{#1}
\csname url@rmstyle\endcsname
\providecommand{\newblock}{\relax}
\providecommand{\bibinfo}[2]{#2}
\providecommand\BIBentrySTDinterwordspacing{\spaceskip=0pt\relax}
\providecommand\BIBentryALTinterwordstretchfactor{4}
\providecommand\BIBentryALTinterwordspacing{\spaceskip=\fontdimen2\font plus
\BIBentryALTinterwordstretchfactor\fontdimen3\font minus
  \fontdimen4\font\relax}
\providecommand\BIBforeignlanguage[2]{{%
\expandafter\ifx\csname l@#1\endcsname\relax
\typeout{** WARNING: IEEEtran.bst: No hyphenation pattern has been}%
\typeout{** loaded for the language `#1'. Using the pattern for}%
\typeout{** the default language instead.}%
\else
\language=\csname l@#1\endcsname
\fi
#2}}

\bibitem{robles2006importance}
G.~Robles-De-La-Torre, ``The importance of the sense of touch in virtual and
  real environments,'' \emph{IEEE Multimedia}, vol.~13, no.~3, pp. 24--30,
  2006.

\bibitem{Jamone2015HighlyHand}
L.~Jamone, L.~Natale, G.~Metta, and G.~Sandini, ``{Highly sensitive soft
  tactile sensors for an anthropomorphic robotic hand},'' \emph{IEEE Sensors
  Journal}, vol.~15, no.~8, pp. 4226--4233, 2015.

\bibitem{LuoShanBimboJoaoDahiyaRavinder2017RoboticReview}
S.~Luo, J.~Bimbo, R.~Dahiya, and H.~Liu, ``Robotic tactile perception of object
  properties: A review,'' \emph{Mechatronics}, vol.~48, pp. 54--67, 2017.

\bibitem{Bartolozzi2016RobotsTouch}
C.~Bartolozzi, L.~Natale, F.~Nori, and G.~Metta, ``Robots with a sense of
  touch,'' \emph{Nature materials}, vol.~15, no.~9, p. 921, 2016.

\bibitem{Gandarias2018EnhancingInteraction}
J.~M. Gandarias, J.~M. G{\'{o}}mez-de Gabriel, and A.~J. Garc{\'{i}}a-Cerezo,
  ``{Enhancing perception with tactile object recognition in adaptive grippers
  for human-robot interaction},'' \emph{Sensors}, vol.~18, no.~3, p. 692, 2018.

\bibitem{Dahiya2010TactileHumanoids}
R.~S. Dahiya, G.~Metta, M.~Valle, and G.~Sandini, ``{Tactile sensing-from
  humans to humanoids},'' \emph{IEEE Transactions on Robotics}, vol.~26, no.~1,
  pp. 1--20, 2010.

\bibitem{Gandarias2018TactileInDisasterScenarios}
J.~M. Gandarias, J.~M. G{\'{o}}mez-de Gabriel, and A.~J. Garc{\'{i}}a-Cerezo,
  ``{Tactile Sensing and Machine Learning for Human and Object Recognition in
  Disaster Scenarios}.''\hskip 1em plus 0.5em minus 0.4em\relax Springer, Cham,
  11 2018, pp. 165--175.

\bibitem{Trujillo-Leon2018TactileWheelchairs}
A.~Trujillo-Leon, W.~Bachta, and F.~Vidal-Verdu, ``{Tactile Sensor Based
  Steering as a Substitute of the Attendant Joystick in Powered Wheelchairs},''
  \emph{IEEE Transactions on Neural Systems and Rehabilitation Engineering},
  vol.~26, no.~7, pp. 1381--1390, 2018.

\bibitem{aggarwal2014object}
A.~Aggarwal and F.~Kirchner, ``Object recognition and localization: The role of
  tactile sensors,'' \emph{Sensors}, vol.~14, no.~2, pp. 3227--3266, 2014.

\bibitem{Kaboli2018Robust}
M.~Kaboli and G.~Cheng, ``Robust tactile descriptors for discriminating objects
  from textural properties via artificial robotic skin,'' \emph{IEEE
  Transactions on Robotics}, vol.~34, no.~4, pp. 985--1003, 2018.

\bibitem{jamali2011majority}
N.~Jamali and C.~Sammut, ``Majority voting: Material classification by tactile
  sensing using surface texture,'' \emph{IEEE Transactions on Robotics},
  vol.~27, no.~3, pp. 508--521, 2011.

\bibitem{liu2012surface}
H.~Liu, X.~Song, J.~Bimbo, L.~Seneviratne, and K.~Althoefer, ``Surface material
  recognition through haptic exploration using an intelligent contact sensing
  finger,'' in \emph{IEEE/RSJ International Conference on Intelligent Robots
  and Systems}, 2012, pp. 52--57.

\bibitem{li2013sensing}
R.~Li and E.~H. Adelson, ``Sensing and recognizing surface textures using a
  gelsight sensor,'' in \emph{IEEE International Conference on Computer Vision
  and Pattern Recognition (CVPR)}, 2013, pp. 1241--1247.

\bibitem{Luo2015NovelRecognition}
S.~Luo, W.~Mou, K.~Althoefer, and H.~Liu, ``{Novel Tactile-SIFT Descriptor for
  Object Shape Recognition},'' \emph{IEEE Sensors Journal}, vol.~15, no.~9, pp.
  5001--5009, 2015.

\bibitem{li2014localization}
R.~Li, R.~Platt, W.~Yuan, A.~ten Pas, N.~Roscup, M.~A. Srinivasan, and
  E.~Adelson, ``Localization and manipulation of small parts using gelsight
  tactile sensing,'' in \emph{IEEE/RSJ International Conference on Intelligent
  Robots and Systems (IROS)}, 2014, pp. 3988--3993.

\bibitem{gandarias2019cnn}
J.~M. Gandarias, A.~J. Garcia-Cerezo, and J.~M. Gomez-de Gabriel, ``{CNN-based
  methods for object recognition with high-resolution tactile sensors},''
  \emph{IEEE Sensors Journal}, vol.~19, no.~16, pp. 6872--6882, 2019.

\bibitem{pastor2019using}
F.~Pastor, J.~M. Gandarias, A.~J. Garc{\'\i}a-Cerezo, and J.~M. G{\'o}mez-de
  Gabriel, ``{Using 3D Convolutional Neural Networks for Tactile Object
  Recognition with Robotic Palpation},'' \emph{Sensors}, vol.~19, no.~24, p.
  5356, 2019.

\bibitem{Madry2014ST-HMP:Data}
M.~Madry, L.~Bo, D.~Kragic, and D.~Fox, ``{ST-HMP: Unsupervised Spatio-Temporal
  feature learning for tactile data},'' in \emph{IEEE International Conference
  on Robotics and Automation (ICRA)}, 2014, pp. 2262--2269.

\bibitem{Liu2016ObjectMethods}
H.~Liu, D.~Guo, and F.~Sun, ``{Object Recognition Using Tactile Measurements:
  Kernel Sparse Coding Methods},'' \emph{IEEE Transactions on Instrumentation
  and Measurement}, vol.~65, no.~3, pp. 656--665, 2016.

\bibitem{gandarias2019active}
J.~M. Gandarias, F.~Pastor, A.~J. Garc{\'\i}a-Cerezo, and J.~M. G{\'o}mez-de
  Gabriel, ``Active tactile recognition of deformable objects with 3d
  convolutional neural networks,'' in \emph{IEEE World Haptics Conference
  (WHC)}.\hskip 1em plus 0.5em minus 0.4em\relax IEEE, 2019, pp. 551--555.

\bibitem{yao2017tactile}
K.~Yao, M.~Kaboli, and G.~Cheng, ``Tactile-based object center of mass
  exploration and discrimination,'' in \emph{IEEE-RAS International Conference
  on Humanoid Robotics (Humanoids)}, 2017, pp. 876--881.

\bibitem{Zapata-Impata2019LearningDetection}
B.~Zapata-Impata, P.~Gil, F.~Torres, B.~S. Zapata-Impata, P.~Gil, and
  F.~Torres, ``{Learning Spatio Temporal Tactile Features with a ConvLSTM for
  the Direction Of Slip Detection},'' \emph{Sensors}, vol.~19, no.~3, p. 523, 1
  2019.

\bibitem{kaboli2016re}
M.~Kaboli, R.~Walker, and G.~Cheng, ``Re-using prior tactile experience by
  robotic hands to discriminate in-hand objects via texture properties,'' in
  \emph{IEEE International Conference on Robotics and Automation (ICRA)}, 2016,
  pp. 2242--2247.

\bibitem{feng2018active}
D.~Feng, M.~Kaboli, and G.~Cheng, ``Active prior tactile knowledge transfer for
  learning tactual properties of new objects,'' \emph{Sensors}, vol.~18, no.~2,
  p. 634, 2018.

\bibitem{kaboli2018active}
M.~Kaboli, D.~Feng, and G.~Cheng, ``Active tactile transfer learning for object
  discrimination in an unstructured environment using multimodal robotic
  skin,'' \emph{International Journal of Humanoid Robotics}, vol.~15, no.~01,
  p. 1850001, 2018.

\bibitem{kroemer2011learning}
O.~Kroemer, C.~H. Lampert, and J.~Peters, ``Learning dynamic tactile sensing
  with robust vision-based training,'' \emph{IEEE Transactions on Robotics},
  vol.~27, no.~3, pp. 545--557, 2011.

\bibitem{Gao2016DeepData}
Y.~Gao, L.~A. Hendricks, K.~J. Kuchenbecker, and T.~Darrell, ``{Deep learning
  for tactile understanding from visual and haptic data},'' in \emph{IEEE
  International Conference on Robotics and Automation (ICRA)}, 6 2016, pp.
  536--543.

\bibitem{Liu2017VisualTactileRecognition}
H.~Liu, Y.~Yu, F.~Sun, and J.~Gu, ``{Visual–Tactile Fusion for Object
  Recognition},'' \emph{IEEE Transactions on Automation Science and
  Engineering}, vol.~14, no.~2, pp. 996--1008, 2017.

\bibitem{Falco2017Cross-modalExploration}
P.~Falco, S.~Lu, A.~Cirillo, C.~Natale, S.~Pirozzi, and D.~Lee, ``{Cross-modal
  visuo-tactile object recognition using robotic active exploration},'' in
  \emph{IEEE International Conference on Robotics and Automation (ICRA)}, 2017,
  pp. 5273--5280.

\bibitem{Pfanne2018FusingEstimation}
M.~Pfanne, M.~Chalon, F.~Stulp, and A.~Albu-Schaffer, ``{Fusing joint
  measurements and visual features for In-Hand object pose estimation},''
  \emph{IEEE Robotics and Automation Letters}, vol.~3, no.~4, pp. 3497--3504,
  2018.

\bibitem{lee2019making}
M.~A. Lee, Y.~Zhu, P.~Zachares, M.~Tan, K.~Srinivasan, S.~Savarese, L.~Fei-Fei,
  A.~Garg, and J.~Bohg, ``Making sense of vision and touch: Learning multimodal
  representations for contact-rich tasks,'' \emph{IEEE Transactions on
  Robotics}, 2020.

\bibitem{xu2013tactile}
D.~Xu, G.~E. Loeb, and J.~A. Fishel, ``Tactile identification of objects using
  bayesian exploration,'' in \emph{IEEE International Conference on Robotics
  and Automation (ICRA)}, 2013, pp. 3056--3061.

\bibitem{kaboli2017tactile}
M.~Kaboli, D.~Feng, K.~Yao, P.~Lanillos, and G.~Cheng, ``A tactile-based
  framework for active object learning and discrimination using multimodal
  robotic skin,'' \emph{IEEE Robotics and Automation Letters}, vol.~2, no.~4,
  pp. 2143--2150, 2017.

\bibitem{GomezEguiluz2018MultimodalSensing}
A.~Gomez~Eguiluz, I.~Rano, S.~Coleman, and T.~McGinnity, ``{Multimodal Material
  Identification through Recursive Tactile Sensing},'' \emph{IEEE Robotics and
  Autonomous Systems}, vol. 106, pp. 130--139, 8 2018.

\bibitem{luo2019iclap}
S.~Luo, W.~Mou, K.~Althoefer, and H.~Liu, ``{iCLAP: Shape recognition by
  combining proprioception and touch sensing},'' \emph{Autonomous Robots},
  vol.~43, no.~4, pp. 993--1004, 2019.

\bibitem{araki2011autonomous}
T.~Araki, T.~Nakamura, T.~Nagai, K.~Funakoshi, M.~Nakano, and N.~Iwahashi,
  ``Autonomous acquisition of multimodal information for online object concept
  formation by robots,'' in \emph{IEEE/RSJ International Conference on
  Intelligent Robots and Systems (IROS)}, 2011, pp. 1540--1547.

\bibitem{navarro2012haptic}
S.~E. Navarro, N.~Gorges, H.~W{\"o}rn, J.~Schill, T.~Asfour, and R.~Dillmann,
  ``Haptic object recognition for multi-fingered robot hands,'' in \emph{IEEE
  haptics symposium (HAPTICS)}, 2012, pp. 497--502.

\bibitem{han2020hands}
M.~Han, S.~Y. G{\"u}nay, G.~Schirner, T.~Pad{\i}r, and D.~Erdo{\u{g}}mu{\c{s}},
  ``{HANDS: a multimodal dataset for modeling toward human grasp intent
  inference in prosthetic hands},'' \emph{Intelligent Service Robotics},
  vol.~13, no.~1, pp. 179--185, 2020.

\bibitem{Library2017OnlineRecognition}
S.~Shahrampour, M.~Noshad, J.~Ding, and V.~Tarokh, ``Online learning for
  multimodal data fusion with application to object recognition,'' \emph{IEEE
  Transactions on Circuits and Systems II: Express Briefs}, vol.~65, no.~9, pp.
  1259--1263, 2017.

\bibitem{chou2018mmdf}
C.-A. Chou, X.~Jin, A.~Mueller, and S.~Ostadabbas, ``{MMDF 2018 Multimodal Data
  Fusion Workshop Report},'' 2018.

\bibitem{medina2019gnss}
D.~Medina, C.~Lass, E.~P{\'e}rez-Marcos, R.~Ziebold, P.~Closas, and
  J.~Garc{\'\i}a, ``{On GNSS Jamming Threat from the Maritime Navigation
  Perspective},'' in \emph{IEEE International Conference on Information Fusion
  (FUSION)}, 2019, pp. 2--5.

\bibitem{fishel2012bayesian}
J.~A. Fishel and G.~E. Loeb, ``Bayesian exploration for intelligent
  identification of textures,'' \emph{Frontiers in neurorobotics}, vol.~6,
  p.~4, 2012.

\bibitem{Schiefer2018ArtificialTasks}
M.~A. Schiefer, E.~L. Graczyk, S.~M. Sidik, D.~W. Tan, and D.~J. Tyler,
  ``Artificial tactile and proprioceptive feedback improves performance and
  confidence on object identification tasks,'' \emph{PloS one}, vol.~13,
  no.~12, 2018.

\bibitem{Okamura2001FeatureFingers}
A.~M. Okamura and M.~R. Cutkosky, ``Feature detection for haptic exploration
  with robotic fingers,'' \emph{The International Journal of Robotics
  Research}, vol.~20, no.~12, pp. 925--938, 2001.

\bibitem{Hochreiter:1997}
S.~Hochreiter and J.~Schmidhuber, ``{Long short-term memory},'' \emph{Neural
  computation}, vol.~9, no.~8, pp. 1735--1780, 1997.

\bibitem{zyner2018recurrent}
A.~Zyner, S.~Worrall, and E.~Nebot, ``A recurrent neural network solution for
  predicting driver intention at unsignalized intersections,'' \emph{IEEE
  Robotics and Automation Letters}, vol.~3, no.~3, pp. 1759--1764, 2018.

\bibitem{park2018multimodal}
D.~Park, Y.~Hoshi, and C.~C. Kemp, ``{A multimodal anomaly detector for
  robot-assisted feeding using an LSTM-based variational autoencoder},''
  \emph{IEEE Robotics and Automation Letters}, vol.~3, no.~3, pp. 1544--1551,
  2018.

\bibitem{shi:2015}
S.~Xingjian, Z.~Chen, H.~Wang, D.-Y. Yeung, W.-K. Wong, and W.-C. Woo,
  ``{Convolutional LSTM network: A machine learning approach for precipitation
  nowcasting},'' in \emph{Advances in neural information processing systems},
  2015, pp. 802--810.

\bibitem{friedman2001elements}
J.~Friedman, T.~Hastie, and R.~Tibshirani, \emph{The elements of statistical
  learning}.\hskip 1em plus 0.5em minus 0.4em\relax Springer series in
  statistics, New York, 2001, vol.~1, no.~10.

\bibitem{arribas2005model}
J.~I. Arribas and J.~Cid-Sueiro, ``A model selection algorithm for a posteriori
  probability estimation with neural networks,'' \emph{IEEE Transactions on
  Neural Networks}, vol.~16, no.~4, pp. 799--809, 2005.

\bibitem{Kerzel}
M.~Kerzel, M.~Ali, H.~G. Ng, and S.~Wermter, ``{Haptic material classification
  with a multi-channel neural network},'' in \emph{International Joint
  Conference on Neural Networks (IJCNN)}, 2017, pp. 439--446.

\bibitem{tatiya2019deep}
G.~Tatiya and J.~Sinapov, ``Deep multi-sensory object category recognition
  using interactive behavioral exploration,'' in \emph{IEEE International
  Conference on Robotics and Automation (ICRA)}, 2019, pp. 7872--7878.

\bibitem{li2020review}
Q.~Li, O.~Kroemer, Z.~Su, F.~F. Veiga, M.~Kaboli, and H.~J. Ritter, ``A review
  of tactile information: Perception and action through touch,'' \emph{IEEE
  Transactions on Robotics}, 2020.

\end{thebibliography}

\addtolength{\textheight}{-20cm}

\end{document}